\documentclass[dvipsnames,format=sigconf,anonymous=false,review=false]{acmart}

\AtBeginDocument{%
  \providecommand\BibTeX{{%
    \normalfont B\kern-0.5em{\scshape i\kern-0.25em b}\kern-0.8em\TeX}}}

\acmConference[Conference'23]{Conference}{}{}
\acmBooktitle{Conference 2023}

\newcommand{\ignore}[1]{}

\usepackage{enumitem}

\usepackage[percent]{overpic}

\begin{document}

\title{ELEA -- Build your own Evolutionary Algorithm in your Browser}

\author{Markus Wagner$^1$, Erik Kohlros$^2$, Gerome Quantmeyer$^2$, Timo Kötzing$^2$}
\affiliation{%
  \institution{$^1$ Monash University, Australia; $^2$ Hasso Plattner Institute/University of Potsdam, Germany}\country{}}
\email{markus.wagner@monash.edu, erik.kohlros/gerome.quantmeyer@student.hpi.uni-potsdam.de, timo.koetzing@hpi.de}

\begin{abstract}
We provide an open source framework to experiment with evolutionary algorithms which we call \emph{Experimenting and Learning toolkit for Evolutionary Algorithms (ELEA)}. ELEA is browser-based and allows to assemble evolutionary algorithms using drag-and-drop, starting from a number of simple pre-designed examples, making the startup costs for employing the toolkit minimal. The designed examples can be executed and collected data can be displayed graphically. Further features include export of algorithm designs and experimental results as well as multi-threading.

With the very intuitive user interface and the short time to get initial experiments going, this tool is especially suitable for explorative analyses of algorithms as well as for the use in classrooms.

\end{abstract}


\maketitle

\section{Introduction}

The \emph{Free Evolutionary Algorithm Kit (FrEAK)} is a ``toolkit to design and analyse evolutionary algorithms, written in Java''~\cite{FrEAK}, developed initially at the University of Dortmund. The last update to the repository was 10 years ago, and technology has evolved significantly since then. With this demonstration paper we introduce a web-based alternative, using the \emph{Blockly} framework developed by Google~\cite{blocklyblogpost,blocklyCode}.

Our intention is not to transfer all and only the functionality of FrEAK to the web, but rather have a completely fresh start, offering functionalities directed at making this tool (1) a starting point for scientific investigations, giving quick insights with little coding overhead; as well as (2) a valuable helper for teaching evolutionary algorithms in the classroom, focusing students' activities on understanding evolutionary algorithms rather than coding them. These two goals aim at making evolutionary algorithms more accessible to a wide range of people, lowering the costs required to get first results. We call our toolkit \emph{Experimenting and Learning toolkit for Evolutionary Algorithms (ELEA)}.

\noindent Specifically, ELEA offers the following features:
\setlist[itemize]{leftmargin=7mm}
\begin{itemize}
    \item Visual Programming System based on Blockly~\cite{blocklyblogpost,blocklyCode}.
    \item Several standard algorithms and test functions implemented.
    \item Export of Code to JavaScript for offline modification and use.
    \item Data visualisation.
    \item Data download as csv or for IOHAnalyzer~\cite{IOHprofiler}.
    \item Multi-threading support.
    \item Open source, can be forked.
\end{itemize}

The visual programming system based on Blockly allows new users to build their own evolutionary algorithms with very little training (see Section~\ref{sec:visualProgrammingSystem} for a discussion on such programming systems). The pre-implemented algorithms and test functions enable first results with the toolkit within minutes. Also, for experienced programmers, using ELEA results in smaller overhead times, because only the EA needs to be designed, and data handling is almost completely taken care of by the system.

In order to allow for far-reaching analyses, the JavaScript sources are available for download, which can then be modified at will for analyses not covered by the functionalities of ELEA. 
As a core feature, ELEA can graphically display data at run time, making the data easily accessible.

We see two main use cases for ELEA, fundamental research and teaching EAs, which we discuss in turn.

\textbf{1. Use for fundamental research.} In the spirit of fast prototyping, ideas or hunches can be fact-checked quickly for small problem sizes using ELEA's existing blocks, but also custom blocks can be designed quickly. Rigorous investigations with multiple repetitions are supported by multi-threading and data management. Furthermore, more complicated examples can be set up by downloading the code and implementing the necessary changes by editing the code directly. We believe that also the run time analysis community can profitably use ELEA.

\textbf{2. Use for teaching EAs.} Teachers and students can explore in-class the effects of algorithmic design decisions. The focus is on designing evolutionary algorithms rather than spending time with data management and program syntax.   Colour-coded blocks assist students to associate colours and shapes~\cite{weintrop2015students} which in turn can help build understanding and retain knowledge. The browser-based environment avoids cross compatibility issues and gives a uniform standard for teachers and students for algorithm definition and benchmarking. At its best, ELEA's building blocks invite students and teachers to ``play'' with algorithmic components, much like with wooden building blocks or Legos, to explore algorithmic spaces in playful and systematic ways. 

You can find a running instance of ELEA as well as its source code at the following URLs:
\begin{center}
    \url{https://elea-toolkit.netlify.app/}\\
    \url{https://github.com/HPI-ELEA/elea}
\end{center}

Another code base that allows to benchmark evolutionary algorithms is the recently developed \emph{Iterative Optimization Heuristics Profiler (IOHprofiler)}~\cite{IOHprofiler}. 
This modular software offers not just evolutionary algorithms, but also EDAs and other optimisation heuristics, as well as a modular approach to the profiling pipeline, including aspects such as automated algorithm configuration. In contrast to the IOHprofiler, ELEA is a lightweight tool based on an easy-to-access graphical user interface. This significantly lowers the initial cost to start using ELEA. In particular, IOHprofiler does not allow for defining algorithms, but a collection of pre-defined algorithms is provided. 
One of the core parts of IOHprofiler is the IOHanalyzer, which offers a wide functionality for investigating the data produced by running algorithms on test functions. This nicely complements ELEA, which can produce compatible data.

There are also programming packages available to run evolutionary algorithms, for example LEAP \cite{LEAP:20}, written in Python. LEAP `` is a general purpose Evolutionary Computation package that combines readable and easy-to-use syntax for search and optimization algorithms with powerful distribution and visualization features.'' While LEAP might be better suited for large scale benchmarking, we believe that the graphical display of the algorithm in ELEA helps understand the underlying principles better than a library call.

The remainder of this paper starts with background on visual programming (in Section~\ref{sec:visualProgrammingSystem}) , proceeds to give an example of using ELEA (in Section~\ref{sec:exampleOfELEA}), before discussing some details of the system design (in Section~\ref{sec:systemDesign}), including multi-threading (see Section~\ref{sec:mt}).

\section{Background on Visual Programming}\label{sec:visualProgrammingSystem}

A visual programming system (VPS)~\cite{wu1999visual} allows users to create programs by manipulating program elements graphically rather than specifying them textually. In a VPS, a user creates a program by arranging ``boxes and arrows'', where boxes represent entities and arrows represent relations.

A VPS can assist programmers to overcome three cognitive challenges~\cite{repenning17lessonslearned}:
\begin{itemize}
    \item Syntactic: arranging programming language components into well-formed programs.
    \item Semantic: assisting users with the comprehension of the meaning of programs.
    \item Pragmatic: bringing a program into a specific situation and understanding its behaviour.
\end{itemize}

Blockly~\cite{blocklyblogpost,blocklyCode} is an example of such a VPS. It is an open-source, client-side library for the programming language JavaScript, providing an editor representing coding concepts as interlocking blocks. Blockly typically runs in a web browser, but it can also generate correct stand-alone code in JavaScript, Python, PHP, Lua, Dart, etc.

\begin{figure}
\centering
\includegraphics[width=\linewidth,trim=0 0 0 0,clip]{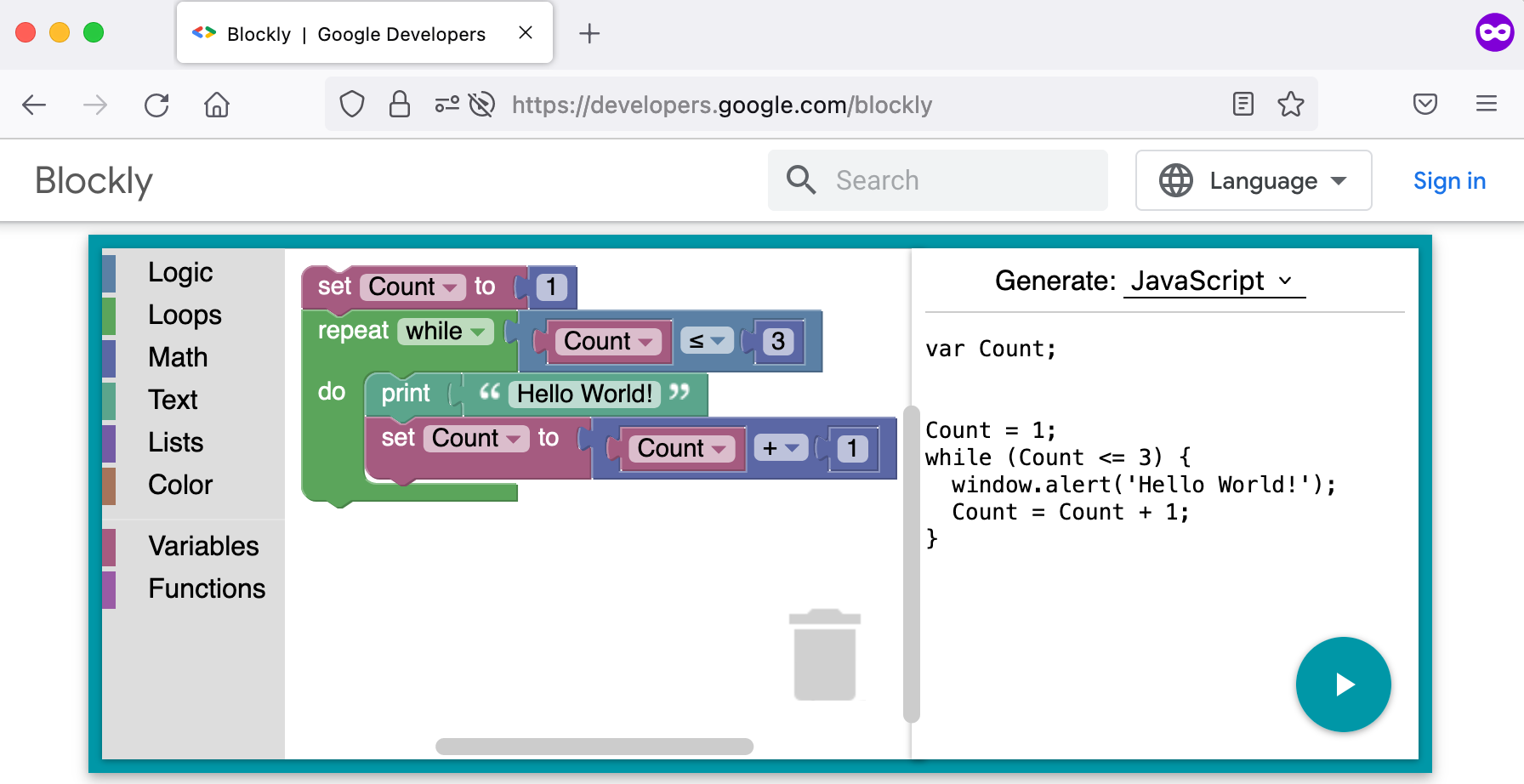}\vspace{-0mm}
\caption{Blockly running in a browser (taken from~\cite{blocklyCode}). 
}
\label{fig:blocklyexample}
\end{figure}

Figure~\ref{fig:blocklyexample} shows an example of Blockly running in a browser. 
The default graphical user interface of the Blockly editor consists of (1) a toolbox, which holds available blocks, and where a user can select blocks; and (2) a workspace, where a user can drag, drop and rearrange blocks. The workspace also includes, by default, zoom icons, and a trash can to delete blocks. Assembly of code consists in drag and drop of functional blocks, giving a final visual impression much like pseudo-code.

For ELEA, we leverage that Blockly is open-source and that custom blocks can be created. Each block consists of a definition, which defines the visual appearance, and a generator, which describes the block's translation to executable code. Blocks can be written in JavaScript, but they can also be defined using blocks.

\section{What's in ELEA?}\label{sec:exampleOfELEA}

\begin{figure*}
\centering

\begin{overpic}[width=\textwidth]{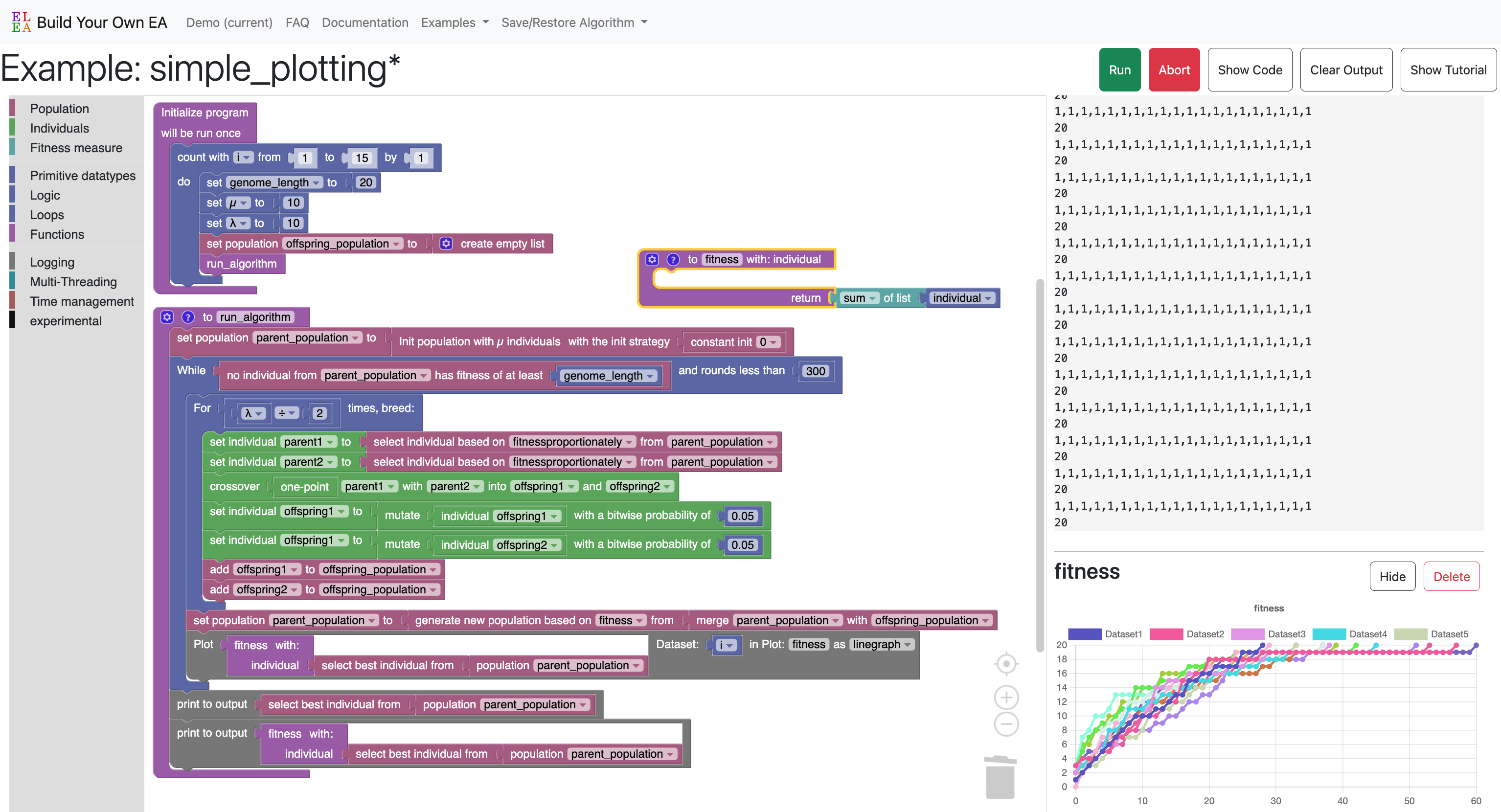}
\put (26,45.5) {\huge \raisebox{.5pt}{\textcircled{\raisebox{-.9pt} {1}}} }
\put (30,33.2) {\huge \raisebox{.5pt}{\textcircled{\raisebox{-.9pt} {2}}} }
\put (50,38.5) {\huge \raisebox{.5pt}{\textcircled{\raisebox{-.9pt} {3}}} }
\put (94,44) {\huge \raisebox{.5pt}{\textcircled{\raisebox{-.9pt} {4}}} }
\put (85,7) {\huge \raisebox{.5pt}{\textcircled{\raisebox{-.9pt} {5}}} }
\end{overpic}
\vspace{-0mm}
\caption{ELEA example ``Simple Plotting''. 
We perform 15 independent repetitions \raisebox{.5pt}{\textcircled{\raisebox{-.9pt} {1}}} of a $(10+10)$-EA that uses one-point crossover and uniform, bit-wise mutation. Individuals are bit strings of length 20. 
The solved problem is the OneMax problem (\raisebox{.5pt}{\textcircled{\raisebox{-.9pt} {3}}}\,: ``sum of list individual''). 
The best individual at the end of each independent run is shown in the top right corner \raisebox{.5pt}{\textcircled{\raisebox{-.9pt} {4}}}, and the quality of the best population member is plotted in the bottom right corner \raisebox{.5pt}{\textcircled{\raisebox{-.9pt} {5}}} (x-axis: generations, y-axis: quality). 
Note: while only a part of the legend is shown, all 15 runs are plotted.}
\label{fig:eleasimpleplotting}
\end{figure*}
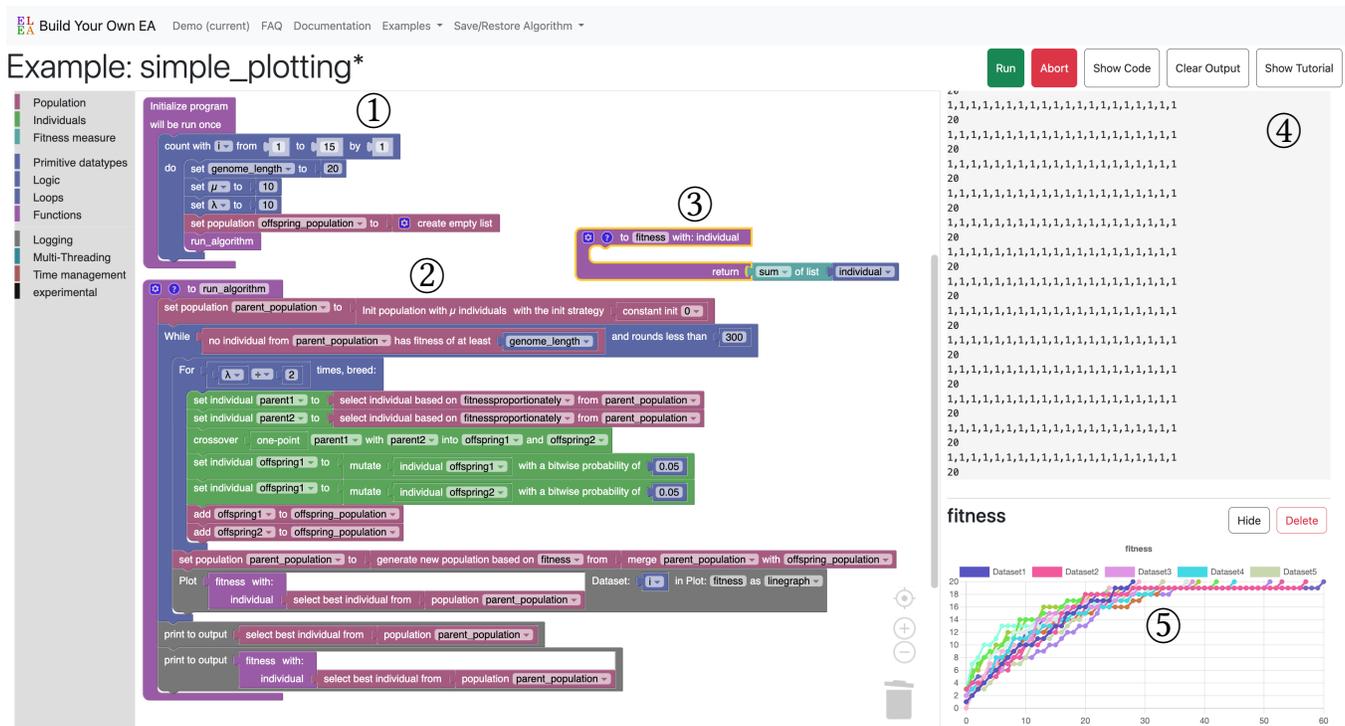

ELEA provides a number of new blocks pertaining to defining evolutionary algorithms, such as blocks for mutation, crossover, selection and so on. 
To demonstrate that it is easy (1) to define an algorithm using ELEA's building blocks and (2) to run it and to plot the results, we show a complete example of a population-based evolutionary algorithm in Figure~\ref{fig:eleasimpleplotting}. 
Details of the particular scenario can be found in the figure's caption.

In the following, we will focus our attention on the coloured building blocks that are used to construct the scenario. 
At present, ELEA provides 11 groups of building blocks, and we outline each group's content next. 
Information on all blocks can be found in ELEA's online documentation. 

\textit{Population (magenta red)} is the largest group of blocks. It provides blocks to initialise a population, such as randomised or explicitly specified. It also provides ways to ``query'' a population, i.e. to acquire information like the size of a population, to get information on the best individual in a population, or to select an individual from a population via methods like uniformly random selection or fitness-proportionate selection. This group also provides blocks to add individuals to a population, to merge or to sort them.
\textit{Individuals (green)} is the second largest group of blocks. They support the initialisation of individuals (randomised or explicitly specified), as well as the crossover (one-point, two-point, uniform) and mutation (mutating each bit with a given probability or mutation a given number of bits).
\textit{Fitness measure (cyan)} provides a number of blocks that calculate fitness functions like OneMax, LeadingOnes, Jump, and some diversity-based metrics. 
\textit{Primitive datatypes (teal-blue)} allow us to create and set variables, to define and concatenate strings, and to create random numbers within a provided range.
\textit{Logic (teal-blue)} allows us to perform conditional executions (akin to an ``if'' in many programming languages) and to assemble Boolean expressions involving Boolean operators including equivalence checks.
\textit{Loops (teal-blue)} enable the simple definition of evolutionary loops, the (number-limited) repetition of selection (e.g. for selecting or creating solutions). It also includes a special block that aggregates data for the later export to IOHAnalyzer.
\textit{Functions (lavender)} comprises special, all-embracing blocks that are needed, for example, to run studies, like the multiple repetition of an algorithm.
\textit{Logging (grey)} provides blocks to print data to the output area of ELEA, to plot data in a scatter/line/bar graph, and to simply add a comment to an algorithm (akin to in-line code comments).
\textit{Multi-Threading (blue cyan)} enables the parallel execution of blocks. As we see this as an important feature, we provide in-depth details in Section~\ref{sec:mt}.
\textit{Time management (copper)} contains blocks for pausing the execution for a number of seconds and for measuring wall-clock time.
\textit{Experimental (black)} covers blocks that do not fit elsewhere. 

All ELEA blocks can be dragged and dropped into place, which lets you quickly assemble an algorithm or modify an existing one. 
Also, blocks can only be connected in particular ways, resulting in code that is always syntactically correct; for example, it is impossible to insert a logging command into a ``parent spot'' of the one-point crossover. 
Lastly, to further increase usability, blocks that are not connected to anything are greyed out, thus making it clear what is executed and what is not.

\section{System Design}\label{sec:systemDesign}

ELEA runs on Node.js and its user interface is built with Bootstrap. The whole tool is based around the Blockly framework, which is developed by Google. Blockly creates an encapsulated module representing the workspace used in our tool and lets us easily build a website around it. In general, it is possible to customise our own blocks using either JavaScript or XML and provide these blocks to the Blockly module. There, we specify how our block should look, meaning for example what parameters can be put in or how the block connects to other blocks, and also provide a snippet of JavaScript code, which is executed in place of the actual block, when the user runs their code inside the tool. This code makes up the algorithm that can be downloaded from the tool. We can also configure how the toolbox looks by adding categories and arranging the blocks in a fitting way. 
Everything else, i.e the moving and combining of blocks, is handled internally by Blockly. For more details see the Google Developers Documentation on Blockly~\cite{blocklyCode}.

Inside Blockly, the algorithm in block-form is represented using XML. This XML can be downloaded and later again uploaded into the workspace. Because every XML element used in this format references a snippet of code, our tool can easily transform this XML data to working JavaScript code.

Since we are using Node.js, you can download your algorithm as JavaScript including a suitable run time environment, and play around without the blocks locally.

\section{Multi-threading}
\label{sec:mt}

To further improve the user experience, ELEA provides multi-threading blocks. Among other, these can be used to efficiently run multiple independent runs of an algorithm, which is a common task undertaken in research on randomised algorithms. 

As an implementation detail, it is important to note that threads in JavaScript have their own scope, which means that they cannot directly use global variables. While this means that worker threads are ``naturally'' separated and thus run independently of each other, it also implies a need to implement (1) information import (e.g.\ to set a mutation rate outside of the actual algorithm, instead of hard-coding it inside), and (2) information sharing back to the main thread (e.g.\ for data logging purposes).

\begin{figure}
\centering
\includegraphics[width=\linewidth,trim=0 0 0 0,clip]{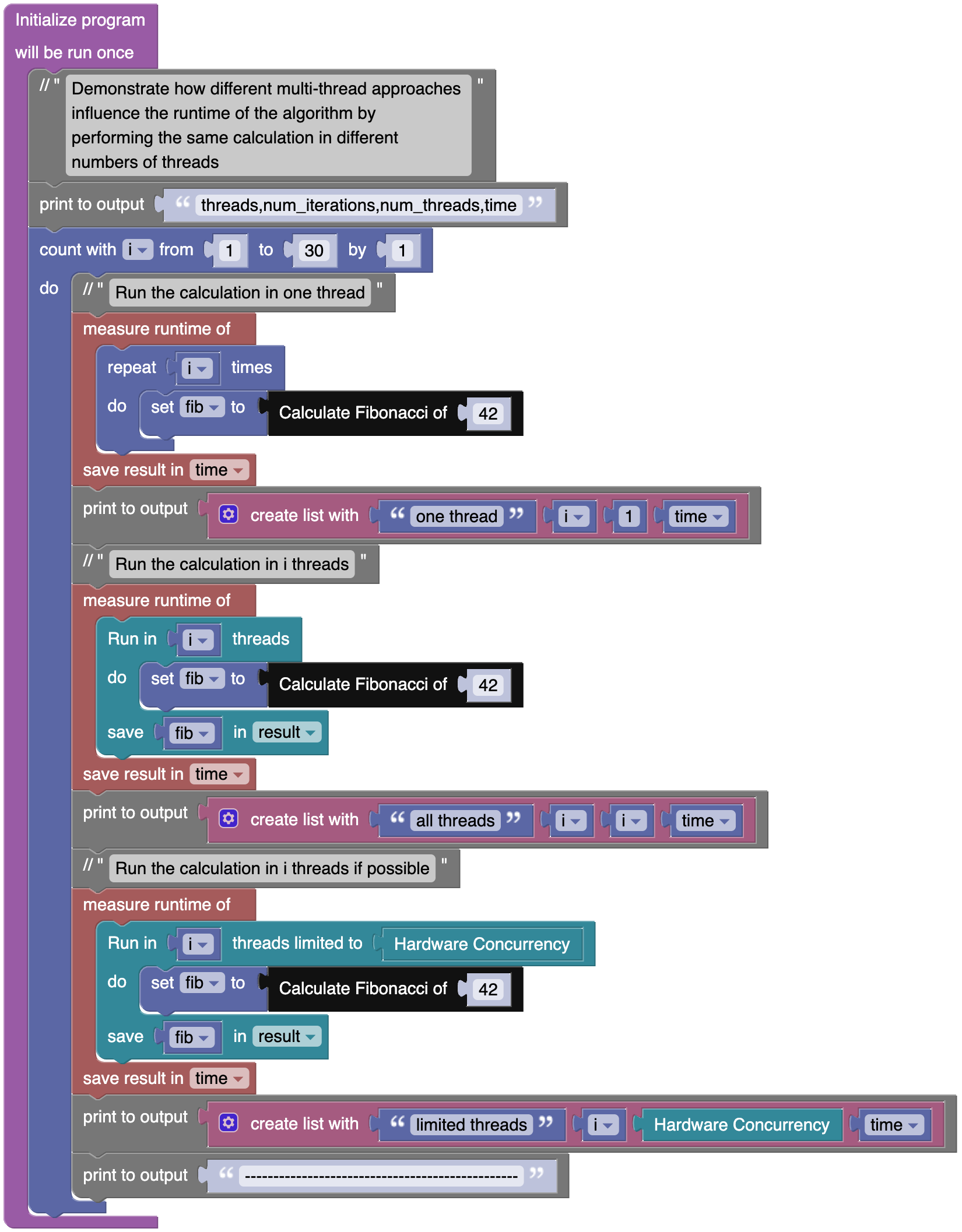}\vspace{-0mm}
\caption{Multi-threading: complete example with a dummy task. The teal-coloured, C-shaped block ``Run in i threads'' are responsible for starting threads and for returning data back to the main thread.}
\label{fig:multithread}
\end{figure}

Figure~\ref{fig:multithread} shows a complete example of multi-threading via our multi-threading blocks; in ELEA, you can find this as the example ``Multi-Threading Performance Test''. 
While this example repeatedly calculates the 42nd Fibonacci number as a dummy task ($1\leq i \leq 30$ repetitions form one task), the online edition of ELEA contains an example using an evolutionary algorithm. 
In particular, this performance test compares three approaches: (1) the sequential execution of $i$ repetitions using one thread, (2) the completely parallel execution (where $i$ threads are started in parallel), and (3) the limited setup, where $x$ parallel threads are used (here: the block ``Hardware Concurrency'' sets $x$ to the number of CPU cores). The last approach can also be limited to, e.g. two worker threads (even when a machine has more cores) to allows the user to retain a responsive computer.

On our computer (a 2022 Macbook Air M2 with 8 CPU cores), the output in ELEA's output window looks like this:
\begin{small}
\begin{verbatim}
  threads,num_iterations,num_threads,time
  one thread,1,1,2136
  all threads,1,1,2045.800000011921
  limited threads,1,8,2140.5
  ------------------------------------------------
  one thread,2,1,4122.900000035763
  all threads,2,2,2139
  limited threads,2,8,2089.7999999523163
  ------------------------------------------------
  one thread,3,1,6250.899999976158
  all threads,3,3,2173.899999976158
  limited threads,3,8,2196.100000023842
  [...]
\end{verbatim}
\end{small}

\begin{figure}
\centering
\includegraphics[width=\linewidth,trim=0 78 0 0,clip]{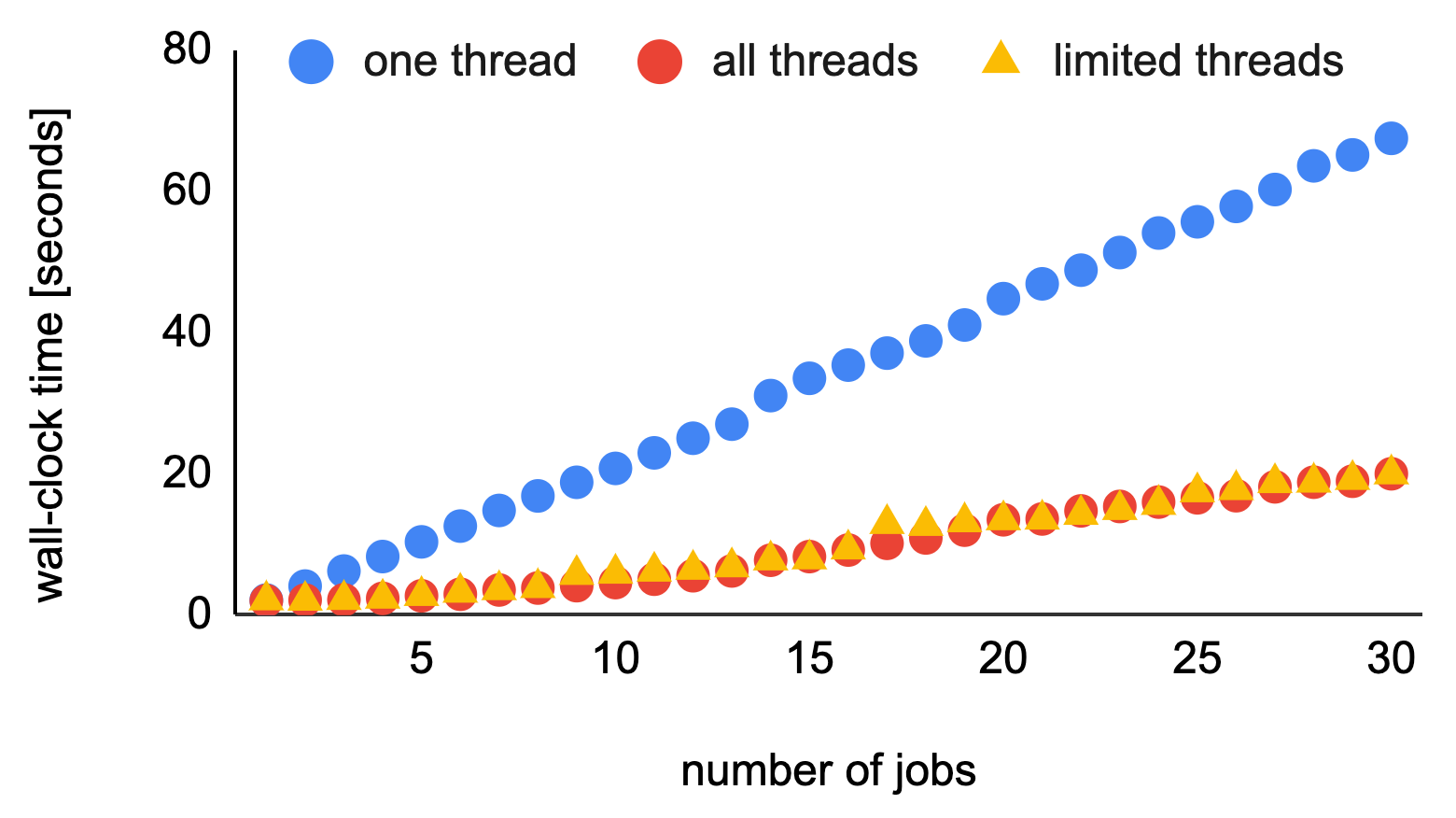}\\
\includegraphics[width=\linewidth,trim=0 0 0 385,clip]{elea-performance.png}\vspace{-0mm}
\caption{Multi-threading: performance results of a simple example. Each data point shows the total wall-clock time of an independent experiment involving the independent calculation of $i \in \{ 1 .. 30 \}$ tasks.}
\label{fig:performance}
\end{figure}

Figure~\ref{fig:performance} shows the results, in particular how the total required time per task changes as the number of $i$ repetitions increases. 
As expected, the wall-clock time for the sequential approach (using  only one thread) increases approximately linearly. 
For the approach where $i$ threads (i.e. up to 30 here) perform the calculations in parallel, run time also increases approximately linearly, although run times are about 70\% faster for large $i$, resulting in a 3.4-fold speedup. 
For the limited approach, we can see ``steps'' in run time increase for the limited approach when multiples of eight are exceeded. 

\begin{acks}
We thank Antony Kamp, Bjarne Sievers, and Oscar Manglaras for their contributions to early versions of ELEA. We thank Xiaoyue Li for her feedback on an earlier version.
\end{acks}

\clearpage

\bibliographystyle{ACM-Reference-Format}
\bibliography{sample-base}


\begin{thebibliography}{8}


\ifx \showCODEN    \undefined \def \showCODEN     #1{\unskip}     \fi
\ifx \showDOI      \undefined \def \showDOI       #1{#1}\fi
\ifx \showISBNx    \undefined \def \showISBNx     #1{\unskip}     \fi
\ifx \showISBNxiii \undefined \def \showISBNxiii  #1{\unskip}     \fi
\ifx \showISSN     \undefined \def \showISSN      #1{\unskip}     \fi
\ifx \showLCCN     \undefined \def \showLCCN      #1{\unskip}     \fi
\ifx \shownote     \undefined \def \shownote      #1{#1}          \fi
\ifx \showarticletitle \undefined \def \showarticletitle #1{#1}   \fi
\ifx \showURL      \undefined \def \showURL       {\relax}        \fi
\providecommand\bibfield[2]{#2}
\providecommand\bibinfo[2]{#2}
\providecommand\natexlab[1]{#1}
\providecommand\showeprint[2][]{arXiv:#2}

\bibitem[Andrea et~al\mbox{.}(2013)]%
        {FrEAK}
\bibfield{author}{\bibinfo{person}{Andrea}, \bibinfo{person}{Dimo Brockhoff},
  \bibinfo{person}{Christian Gunia}, \bibinfo{person}{Matthias Englert},
  \bibinfo{person}{Oliver Heering}, \bibinfo{person}{Michael Leifhelm},
  \bibinfo{person}{Heiko R{\"o}glin}, \bibinfo{person}{Patrick Briest},
  \bibinfo{person}{Dirk Sudholt}, \bibinfo{person}{Stefan Tannenbaum}, {and}
  \bibinfo{person}{Thomas Jansen}.} \bibinfo{year}{2013}\natexlab{}.
\newblock \bibinfo{title}{Free Evolutionary Algorithm Kit (FrEAK)}.
\newblock
\newblock
\urldef\tempurl%
\url{https://sourceforge.net/projects/freak427/}
\showURL{%
Retrieved January 31, 2023 from \tempurl}


\bibitem[Black(2012)]%
        {blocklyblogpost}
\bibfield{author}{\bibinfo{person}{Lucy Black}.}
  \bibinfo{year}{2012}\natexlab{}.
\newblock \bibinfo{title}{Google Blockly -- A Graphical Language with a
  Difference}.
\newblock
\newblock
\urldef\tempurl%
\url{https://www.i-programmer.info/news/98-languages/4357-google-blockly-a-graphical-language-with-a-difference.html}
\showURL{%
Retrieved January 31, 2023 from \tempurl}


\bibitem[Coletti et~al\mbox{.}(2020)]%
        {LEAP:20}
\bibfield{author}{\bibinfo{person}{Mark~A. Coletti}, \bibinfo{person}{Eric~O.
  Scott}, {and} \bibinfo{person}{Jeffrey~K. Bassett}.}
  \bibinfo{year}{2020}\natexlab{}.
\newblock \showarticletitle{Library for Evolutionary Algorithms in Python
  (LEAP)}. In \bibinfo{booktitle}{\emph{Proc.\ of GECCO'20}}.
  \bibinfo{publisher}{ACM}, \bibinfo{pages}{1571--1579}.
\newblock
\showISBNx{9781450371278}
\urldef\tempurl%
\url{https://doi.org/10.1145/3377929.3398147}
\showDOI{\tempurl}


\bibitem[Doerr et~al\mbox{.}(2018)]%
        {IOHprofiler}
\bibfield{author}{\bibinfo{person}{Carola Doerr}, \bibinfo{person}{Hao Wang},
  \bibinfo{person}{Furong Ye}, \bibinfo{person}{Sander van Rijn}, {and}
  \bibinfo{person}{Thomas B{\"{a}}ck}.} \bibinfo{year}{2018}\natexlab{}.
\newblock \showarticletitle{IOHprofiler: {A} Benchmarking and Profiling Tool
  for Iterative Optimization Heuristics}.
\newblock \bibinfo{journal}{\emph{CoRR}}  \bibinfo{volume}{abs/1810.05281}
  (\bibinfo{year}{2018}).
\newblock
\urldef\tempurl%
\url{http://arxiv.org/abs/1810.05281}
\showURL{%
\tempurl}


\bibitem[Google(2023)]%
        {blocklyCode}
\bibfield{author}{\bibinfo{person}{Google}.} \bibinfo{year}{2023}\natexlab{}.
\newblock \bibinfo{title}{Blockly landing page}.
\newblock
\newblock
\urldef\tempurl%
\url{https://developers.google.com/blockly/}
\showURL{%
Retrieved January 31, 2023 from \tempurl}


\bibitem[Repenning(2017)]%
        {repenning17lessonslearned}
\bibfield{author}{\bibinfo{person}{Alexander Repenning}.}
  \bibinfo{year}{2017}\natexlab{}.
\newblock \showarticletitle{Moving Beyond Syntax: Lessons from 20 Years of
  Blocks Programing in AgentSheets}.
\newblock \bibinfo{journal}{\emph{Journal of Visual Lang.\ and Computing}}
  \bibinfo{volume}{3} (\bibinfo{year}{2017}), \bibinfo{pages}{68--91}.
\newblock
\urldef\tempurl%
\url{http://ksiresearch.org/vlss/journal/VLSS2017/vlss-2017-repenning.pdf}
\showURL{%
\tempurl}


\bibitem[Weintrop and Wilensky(2015)]%
        {weintrop2015students}
\bibfield{author}{\bibinfo{person}{David Weintrop} {and} \bibinfo{person}{Uri
  Wilensky}.} \bibinfo{year}{2015}\natexlab{}.
\newblock \showarticletitle{To Block or Not to Block, That is the Question:
  Students' Perceptions of Blocks-Based Programming}. In
  \bibinfo{booktitle}{\emph{Proc.\ of IDC'15}}. \bibinfo{publisher}{ACM},
  \bibinfo{address}{New York, NY, USA}, \bibinfo{pages}{199–208}.
\newblock
\showISBNx{9781450335904}
\urldef\tempurl%
\url{https://doi.org/10.1145/2771839.2771860}
\showDOI{\tempurl}


\bibitem[Wu et~al\mbox{.}(1999)]%
        {wu1999visual}
\bibfield{author}{\bibinfo{person}{Annie~S Wu}, \bibinfo{person}{Kenneth~A
  De~Jong}, \bibinfo{person}{Donald~S Burke}, \bibinfo{person}{John~J
  Grefenstette}, {and} \bibinfo{person}{C~Loggia Ramsey}.}
  \bibinfo{year}{1999}\natexlab{}.
\newblock \showarticletitle{Visual analysis of evolutionary algorithms}. In
  \bibinfo{booktitle}{\emph{IEEE Congress on Evolutionary Computation}}. IEEE,
  \bibinfo{pages}{1419--1425}.
\newblock


\end{thebibliography}

\end{document}